# Resolving Conflicting Arguments under Uncertainties

**Benson Hin-Kwong Ng, Kam-Fai Wong, Boon-Toh Low**
Department of Systems Engineering & Engineering Management
The Chinese University of Hong Kong
{hkng, kfwong, btlow}@se.cuhk.edu.hk

## Abstract

Distributed knowledge based applications in open domain rely on common sense information which is bound to be uncertain and incomplete. To draw the useful conclusions from ambiguous data, one must address uncertainties and conflicts incurred in a holistic view. No integrated frameworks are viable without an in-depth analysis of conflicts incurred by uncertainties. In this paper, we give such an analysis and based on the result, propose an integrated framework. Our framework extends definite argumentation theory to model uncertainty. It supports three views over conflicting and uncertain knowledge. Thus, knowledge engineers can draw different conclusions depending on the application context (i.e. view). We also give an illustrative example on strategical decision support to show the practical usefulness of our framework.

## 1 INTRODUCTION

Expert systems rely on common sense knowledge which is conflicting and uncertain. Conflicts and uncertainties are not orthogonal but intertwined. Uncertainties usually lead to conflicts and conflicts usually result in uncertainties. Thus, any practical expert systems must address both aspects.

Logic-based information reasoning is an interesting domain where conflicts, in the form of partial and opposing arguments, are unavoidable. With an eye on this particular domain, we propose a framework to capture the notion of conflict and resolve it under uncertainties.

The most widely used methodology for tackling conflicts and uncertainties are model-based approaches like classical logic. Owing to their declarative nature, they have difficulties in discerning different kinds of conflicts. On the other hand, argumentation, a reviving integrated approach, gives promising results addressing this issue. It studies not only models of knowledge bases but also their structures, derivations and conflicts. Unifying these studies can lead to a deeper understanding of uncertain and conflicting knowledge bases.

John Pollock pioneered the research of defeasible argumentation (Pollock, 1990) but he only concentrated on probability analysis of the topic. P. M. Dung (Dung, 1995b) and Robert Kowalski (Kowalski and Toni, 1996) investigated how argumentation subsume other non-monotonic logics and logic programming frameworks with minimalistic settings. Gerard Vreeswijk (Vreeswijk, 1997) employed argumentation to model multi-agent systems in a completely abstract way. Henry Prakken shed light on argumentation for pragmatic legal applications with his extended logic programming framework. Bart Verheij (Verheij, 1996) unified all contemporary argumentation within his two frameworks. Our framework differs from these researches in that we concentrate on conflicts and uncertainties in a distributed setting.

The rest of the paper is organized as follows: In Section 2, we describe our approach of modeling uncertainties in argumentation and the conflicts that may pertain to such a setting. After introducing various definitions in Section 3, Section 4 and 5 outline the conflicts and possible resolution methods, respectively. We proceed to semantical analysis in Section 6 and dialectical analysis of proof-theory in Section 7. In Section 8, we describe how our framework relates to existing ones. In Section 9, a typical application of our framework is outlined. In the conclusion, Section 10, we discuss shortcomings of our framework and propose future work.

## 2  MODELING UNCERTAINTIES IN ARGUMENTATION

### 2.1  Background

Uncertainty takes many forms. In many application domains, quantitative measures of uncertainty are difficult to obtain (Fox et al., ; Paul Krause and Fox, 1995), e.g. logic based information retrieval, rule based analysis, etc. In this paper, we focus on qualitative aspects of uncertainty. Undoubtedly, logic is one of the most powerful qualitative approaches. In classical model-based analysis of logic, entailment relation is declarative. Inference structure is not important as long as a fact can be derived. This is clearly undesirable for uncertainty reasoning. To draw useful conclusions from a set of potentially conflicting and uncertain knowledge, we must know the structural aspect of inferencing. We must know in what way conclusions are drawn and at which step conflicts are introduced.

Argumentation is a variant of logic which can attack conflicts from their structures. Briefly, it was revived by Toulmin's book (Toulmin, 1958). Since then, many potential applications of the theory in nonmonotonic reasoning have been suggested (Lin, 1993; Lin and Shoham, ). The methodology of argumentation shows great similarity to the logic programming paradigm. As a result, various attempts (Dung, 1995a; Dung, 1995b; Prakken and Sartor, 1997) have been made on blending the practical usefulness of logic programming and the philosophical insight of argumentation.

In this paper, we show that argumentation is also applicable and, in fact, well-suited to the problem of uncertainty reasoning. Inspired by the work of Dung (Dung, 1995b) and Prakken (Prakken and Sartor, 1997) , our framework is based on logic programming. We represent knowledge with Extended Disjunctive Logic Program ($EDLP$)(Gelfond and Lifschitz, 1991) which is an extension of their framework. We introduce disjunctive in the head of a clause for representing uncertain information. For example, $\{r_1 : dog\_bark \rightarrow stranger \vee arson\}$ means *"a dog barks if either it sees a stranger or an arson"*.

### 2.2  Our Framework

Our argumentation framework assumes a set of distributed knowledge based systems ($KBSs$) (or agents). Conflicts are divided into two levels, namely

1. Intra-KBS-conflicts - conflicts within one $KBS$

2. Inter-KBS-conflicts - conflicts between $KBSs$

The network of KBSs are managed by parties with different interests, such as different departments in a university. Knowledge in the KBSs overlap with each other either entirely or partially. This overlapping will most likely contain deviant views to the same knowledge. For example, in the government organization of a country, the industrial development council is usually pessimistic to possible outcome of raising interest-rate whereas the banking authority council usually bears an opposite view, viz:

- **Industry** : raising-interest-rate *may lead to* pressure-on-industry

- **Banking** : raising-interest-rate *may lead to* pressure-on-industry or stimulate-financial-related-sectors

To be general, one would like to keep multiple perspectives of the two views and apply them according to context.

The problem of reasoning in a distributed setting is then not to change either KBSs as other parts of them rely on their own views. The objective is therefore to recognize differences between different views/opinions yet still enables knowledge engineers to draw useful conclusions out of them. The discrepancies between distributed KBSs may lead to non-logical contradiction, and this is very difficult to model simply by classical logic.

## 3  DEFINITION

Formally, our argumentation framework is defined, in a top-down fashion, as follows: If a literal $l$ is an atom $a$ then $\bar{l}$ is $\neg a$. If $l$ is $\neg a$ then $\bar{l}$ is $a$. An argumentation system $AS$ is a binary-tuple $< Ags, Pg >$ where $Ags$ is a collection of distributed argumentation agents and $Pg$ is a preference hierarchy between these agents. An argumentation agent $Ag$ is a binary-tuple $< R, P >$ where $R$ is a set of rules and $P$ is a preference hierarchy between these rules. A preference hierarchy $P$ is a set of ordered pair in which we say "$s$ is preferred than $g$" (denoted by $s > g$) if and only if $< s, g >\in P$. A rule is an $EDLP$ clause of the form,

$$r : a_1 \wedge \ldots \wedge a_l \wedge \sim a_{l+1} \wedge \ldots \wedge \sim a_m \rightarrow a_{m+1} \vee \ldots \vee a_n$$

where $r$ is the name of the rule, $a_1, \ldots, a_n$ are all literals, $\sim$ is the non-provable sign[1]. We denote $Cd(r) = \{a_1, \ldots, a_m\}$, $Cn(r) = \{a_{m+1}, \ldots, a_n\}$,

---

[1] $\sim a$ means $a$ is not derivable.

$Strong(Cd(r)) = \{a_1, \ldots, a_l\}$ and $Weak(Cd(r)) = \{a_{l+1}, \ldots, a_m\}$. An argument $Arg$ is a finite sequence of rules $\{r_0, \ldots, r_N\}$ in which every rule $r_i$ satisfies the following conditions:

1. If $p \in Strong(Cd(r_i))$ then there exists $r \in \{r_0, \ldots, r_{i-1}\}$ and $p = certain(r)$.

2. There exists $q \in Cn(r_i)$ and denoted as $certain(r_i)$, such that if $l \in \{Cn(r_i) - q\}$ then $\bar{l} \in \bigcup\limits_{j=0}^{i-1} certain(r_j)$.

3. There does not exists $r \in \{r_0, \ldots, r_{i-1}\}$ such that $certain(r_i) = certain(r)$ and $certain(r_i) = \overline{certaih(r)}$.

The function $certain$ denotes a "definite" reading of an "indefinite" clause. $certain(r)$ denotes a literal in the conclusion of an indefinite rule such that all other literals are pruned by their complementary counterparts other rules' $certain$. For example, the value of $certain(r_2)$ is $a$ in the following argument: $\{r_1 :\rightarrow \neg b, r_2 :\rightarrow a \vee b\}$. Condition 1 ensures that every rule in an argument must be grounded. Conditions 2 and 3 ensure that a rule must have a unique meaning and rules in argument are not redundant, respectively. Restriction imposed by condition 2 is for the sake of simplicity in demonstrating inter-KBS-conflicts.

## 4    CONFLICTS WITHIN A $KBS$

Under our framework, conflicts within a $KBS$ is referred to as intra-KBS-conflicts. As an example, consider the following segment of a legal knowledge base $< R, P >$.

$r_4 : \quad \rightarrow finger\_print$
$r_5 : \quad finger\_print \rightarrow murderer \vee owner$
$r_7 : \quad \sim ownership \rightarrow \neg owner$
$r_8 : \quad \sim criminal\_record \rightarrow \neg murderer$
$r_6 : \quad murderer \rightarrow put\_into\_jail$
$r_9 : \quad \rightarrow criminal\_record$

From the above, the following arguments are found:

1. $Arg_1$ is comprised of $r_4, r_7$, and $r_5$.

2. $Arg_2$ entails $Arg_1$ and $r_6$. Irrespective of the outcome of the proof of $ownership$ ($r_7$), $Arg_2$ always leads to the conclusion that the $murderer$ should be $put\_into\_jail$. Thus, both $murderer$ and $put\_into\_jail$ are $certain()$ in $Arg_1$.

3. $Arg_3$ is formed solely by $r_8$. Thus, $certain()$ of $Arg_3$ (i.e. $\neg murderer$) is inconsistent with that of $Arg_2$ (i.e. $murderer$). This kind of head-on conflicts is called rebut.

4. $Arg_4$ entails $r_9$ alone. Note that it attacks the assumption of $Arg_3$. We call this undercut.

From the above example, we define the following:

**Definition 1** $Arg$ under-cuts $Arg'$ if and only if there exists a rule $r$ in $Arg$, a rule $r'$ in $Arg'$ and a literal $p$ in $Weak(Cd(r'))$ such that $p = certain(r)$.

**Definition 2** $Arg$ rebut $Arg'$ if and only if there exists a rule $r'$ in $Arg'$ and a rule $r$ in $Arg$ satisfying $certain(r) = certain(r')$, $r' > r \notin P$.

**Definition 3** $Arg$ defeats $Arg'$ if,

1. $Arg$ is empty and, $Arg'$ undercuts itself; or

2. $Arg$ undercuts $Arg'$; or

3. $Arg$ rebuts $Arg'$ and $Arg'$ does not undercut $Arg$.

Then, we define the asymmetric order *strictly defeat* as below.

**Definition 4** $Arg$ strictly defeats $Arg'$ if only if $Arg$ defeats $Arg'$ but not $Arg'$ defeats $Arg$.

## 5    CONFLICTS BETWEEN $KBSS$

In this section, we consider conflicts aroused in an argumentation system $< Ags, Pg >$. Conflicts pertain to a single $KBS$ can also happen in a group of $KBS$s. To reflect that $rebut$ is based on preference hierarchy, we add the following to the previous definition 2.

**Definition 5** $Arg$ of $Ag_1 \in Ags$ rebuts $Arg'$ of $Ag_2 \in Ags$ if and only if there exists a rule $r_1$ in $Arg$ and a rule $r_2$ in $Arg'$ such that $certain(r_1) = certain(r_2)$ and $Ag_2 > Ag_1 \notin Pg$.

Next, we shall consider conflicts pertaining to uncertainties between different $KBS$s. When knowledge is distributed over a set of agents, it is not uncommon to see that the same piece of knowledge represented differently in different agents. For example, Table 1 shows two agents, $Agt_1$ and $Agt_2$, which possess two variants of the same knowledge. It is intuitive to see that $Agt_1$ will conclude $put\_into\_jail$ whereas $Agt_2$ will not, owing to the uncertainty in $r_5$. A direct merge of $Agt_1$ and $Agt_2$ will lead to $put\_into\_jail$ as $Agt_1$ dominates the union in the sense of classical logic. However, such conclusion essentially assumes certain information with higher preference than uncertain information and is credulous in this regard. It is arguable that on what basis can we bear this assumption and cherish certain information rather than uncertain one? And on what basis we want to do it the other way round?

Table 1: Example of Two Distributed KBSs

| AGENT | KNOWLEDGE |
|-------|-----------|
| $Agt_1$ | $r_1 :\to finger\_print$ |
| | $r_2 : finger\_print \to murderer$ |
| | $r_3 :\sim murderer \to release$ |
| | $r_6 : murderer \to put\_into\_jail$ |
| | |
| $Agt_2$ | $r_4 :\to finger\_print$ |
| | $r_5 : finger\_print \to murderer \vee owner$ |
| | $r_6 : murderer \to put\_into\_jail$ |

## 5.1 Credulous View

From the viewpoint of classical logic, we can interpret a rule $r_2$ as a revision of another rule $r_5$ if one shows that $r_5$ is too loose a statement (see Table 1). This is exactly what embedded in the essence of classical logic. We called this a credulous view of distributed knowledge.

In many traditional reasoning frameworks, without the non-provability sign $\sim$, credulous view can help to enlarge the set of positive information. However, this is not the case for our framework and most other non-monotonic systems which attack default reasoning. This can be shown by the following example which consists of two agents $Agt_1$ and $Agt_2$ (see Table 1).

If we adopt the approach of classical logic, $Agt_1$ will clearly out-rate $Agt_2$. Further, $r_3$ is not applicable and hence, *release* is not deduced (which is deducible under skeptical view, defined in the next section). Thus, credulous view is not necessarily additive in the presence of a skeptical view.

## 5.2 Skeptical View

When our multi-agent framework is applied to the legal reasoning domain, credulous inference is highly undesirable due to the spirit of "proof beyond any reasonable doubts". Consider the last example, if ownership is also a plausible explanation for the evidence of one's finger print on an object, then how can we arrive at the conclusion of *murderer* with certainty? In fact, it is the job of the defense legal agent, in practice, to retrieve as many uncertain information as possible from legal knowledge bases and use them to presents "reasonable doubts". Under such circumstances, we would prefer $Agt_2$'s view to $Agt_1$. Such preference is actually a form of suppression and reveals conflicts between different $KBSs$. Notice that this type of conflict aroused because of the great similarity between two variants of the same piece of information being distributed. In

syntactic form, we introduce

**Definition 6** A rule $r_1$ of $Agt_1$ thins a rule $r_2$ of $Agt_2$ if and only if $Cd(r_2) = Cd(r_1)$ and $Cn(r_2) \subset Cn(r_1)$.

The semantic of thinning is self-explanatory. $r_1$ is a more general statement about the relation between $Cd(r_2)$ and $Cn(r_2)$. It says that $Cn(r_1) - Cn(r_2)$ are also possible outcomes of $Cd(r_2)$. In a modal sense, the necessity implied $r_1$ is incorrect. Semantically, $r_2$ in $Agt_2$ weakens the certainty of $r_1$ in $Agt_1$. We call this type of uncertain conflicts *thinning*. It happens due to a skeptical view in the distributed knowledge environment. In general, we refer this as inter-KBS-conflicts.

Inter-KBS-conflicts pertain to any inference schema which supports qualitative uncertainties like disjunctive in a distributed knowledge setting. A straight forward resolution scheme to inter-KBS-conflicts is to eliminate the additional uncertainties introduced, i.e. $Cn(r_1) - Cn(r_2)$.

For intra-KBS-conflicts, rules related to the judgment is easily defined. It is, however, not the same for inter-KBS-conflicts. As inter-KBS-conflicts aroused because of the additional uncertainties in a rule, we consider a set of auxiliary rules to determine the defeating status

1. If a rule $r_1$ thins a rule $r_2$, the set of auxiliary rules is $auxiliary(r_1, r_2) = \{Cd(r_1) \to q | q \in (Cn(r_1) - Cn(r_2))\}$ and every rule is an alias of $r_1$ in preference hierarchy.

2. An argument $Arg$ is not defeated by a rule $r_1$ if $r_1$ thins a rule $r_2$ in $Arg$ and for every auxiliary rule $r$ in $auxiliary(r_1, r_2)$, there exists an justified argument defeating it.

To illustrate the idea here, let us consider the example in Table 1 again. There is an argument $Arg_1 = \{r_1, r_2, r_6\}$ in $Agt_1$. By definition, $r_5$ and $r_2$ of $Arg_1$ show inter-KBS-conflict. The set of auxiliary rules is defined as,

$$auxiliary(r_5, r_2)$$
$$= \{finger\_print \to q | q \in (owner)\}$$
$$= \{finger\_print \to owner\}$$

To maintain $Arg_1$'s undefeated status, additional arguments must be found to defeat auxiliary rules in $auxiliary(r_5, r_2)$. However, there is no such a rule in $Agt_1 \cup Agt_2$.

Similar to credulous view, skeptical view also does not give us more certain information in reasoning. Consideration of the simple example in the last subsection helps to illustrate this behavior. If we consider $r_5$ as a thinning attacker, *put_into_jail* cannot be derived from $Agt_1$. Thus, a skeptical view does not, in general, enlarge the set of facts concluded.

### 5.3  Generalized Skeptical View

In a skeptical view, we only focus on a particular type of similarity between disjunctive clauses. In this section, we extend the analysis to general similarity between two similar but yet distributed disjunctive clauses. In Table 2, we show the generalization of conflicting clauses of this kind. In the table , we assume that both $R_A, R_B$ and $R_C, R_D$ are from different KBSs. *Cond* denotes the conjunctive condition where as $\Gamma$, $\Delta$ and *Conc* denote the disjunctive literal sequents. Also, $R_A$ and $R_C$ are parts of arguments $Arg_A$ and $Arg_C$, respectively.

Table 2: Conflicts Analysis Between Similar Clauses

| TYPE | SCENARIO |
|---|---|
| Subsumption | $R_A : Cond \Rightarrow Conc$ |
| | $R_B : Cond \Rightarrow Conc \vee \Delta$ |
| | |
| Intersection | $R_C : Cond \Rightarrow Conc \vee \Gamma$ |
| | $R_D : Cond \Rightarrow Conc \vee \Delta$ |

Two rules $R_A$ and $R_B$ are similar if and only if

1. Conditions of $R_A$ and $R_B$ are the same; and

2. Conclusion of $R_A$ intersects with that of $R_B$.

The notion of similarity is an extension of the skeptical view in which only subsumption is considered. In this extended setting, $R_C$ and $R_D$ are conflicting and $R_D$ thins $Arg_C$. To determine the result, we consider the status of the rules involved, i.e. the unique meaning of $R_C$ in $Arg_C$ .

In both situations, we show that a generalized skeptical view does not introduce new things and can be tackled by the following techniques:

1. Case 1 : If $certain(R_C) \in Conc$,
   - There exists arguments defeating $Cn(R_C) - certain(R_C)$.
   - $\Gamma \subseteq \{Cn(R_C) - certain(R_C)\}$.
   - Thus, there exists arguments defeating $\Gamma$.
   - $R_C$ degenerates to $R'_C : Cond \Rightarrow Conc$.
   - The problem is then reduced to subsumption problem in previous section
     - $R'_C : Cond \Rightarrow Conc$

   - $R_D : Cond \Rightarrow Conc \vee \Delta$

2. Case 2 : If $certain(R_C) \in \Gamma$,
   - There exists arguments defeating *Conc*.
   - The problem degenerates into
     - $R'_C : Cond \Rightarrow \Gamma$
     - $R_D : Cond \Rightarrow Conc \vee \Delta$
   - The conflict criteria are no longer met and thus can be neglect.

The simplicity achieved in Case 2 is due to our unique meaning restriction imposed on the argument definition. The situation will be extremely complex when the restriction is relaxed.

## 6  SEMANTICS

Now we have the bells and whistles to define our argumentation semantics for resolving conflicts under uncertainties. Formally, our semantics is based on a fix-point operator $\Pi$ which operates on two sets of arguments $ArgSet$ and $S$. $\Pi_S(ArgSet)$ gives a subset of $S$ such that all their counter-arguments/defeaters are strictly defeated by arguments in $Arg$. It can be proved that $\Pi$ is monotone in credulous view, skeptical view and generalized skeptical view. The fix-point operator $\Pi$ is essentially the same as Prakken's (Prakken and Sartor, 1997). Our semantics differ from his in the definition of "strictly defeat" which is the core concept of argumentation. Indeed, there are three kinds of fix-point operator. The key point is in what way the notion of counter-argument is interpreted. Table 3 summarizes the notion of counter-argument for different views shown in Sections 5.1 and 5.2.

Table 3: Counter-Arguments For Different Views

| | CREDULOUS | SKEPTICAL |
|---|---|---|
| COUNTER-ARGUMENTS | self-defeat | self-defeat |
| | undercut | undercut |
| | rebut | rebut |
| | | *thinning* |

To achieve monotonicity, the fix-point operator $\Pi$ relies on the notions of asymmetric order ( "*strictly defeat*") which we introduced in the previous sections.

By *Knaster-Tarski* theorem, the fix-point $\Pi^*$ of $\Pi$ with respect to a set of argument $S$ exists and can be obtained as below.

$$F(0) = \Pi_S(\emptyset)$$
$$F(i) = \Pi_S(F(i-1))$$
$$\Pi_S^* = \lim_{i \to \infty} F(i)$$

Let $ArgSet$ be the set of all arguments which can be constructed from an argumentation system $AS$. An argument $Arg$ is *justified* with respect to $AS$ if and only if $Arg \in \Pi^*_{ArgSet}$; it is *defeated* if and only if there exists a justified argument $Arg' \in AS$ strictly defeating it; and it is *arguable*, otherwise.

## 7   DIALECTICAL PROOF THEORY

Dung proposed a dialectical proof theory for argumentation in (Dung, 1995b). Prakken adopted it in his fixed priority framework. As the proof-theory is defined on arguments, we show that a simplified version of it can also be used in our framework.

A proof of an argument is defined on an argument tree. Each internal/external node is an argument and their child nodes are their defeaters. An argument tree $T$ is constructed as below.

1. Level 1 is the proposition to be justified.

2. At an odd (even) level, a proponent (opponent) makes moves to strictly defeat all (any) moves from opponents (proponents) at previous levels.

3. In any branch, an opponent cannot make the same move twice.

A proponent is said to win a branch of an argument tree if and only if the opponent cannot make any further move at the branch; and said to win an argument tree if and only if it wins all branches. An argument $Arg$ is *provably justified* if it wins an argument tree at level 1; it is *provably defeated* if it is defeated by a provably justified argument; and it is *provably arguable*, otherwise. It can be shown that an argument $Arg$ is provably justified with respect to an argumentation system $AS$ if and only if it is justified with respect to $AS$ [2].

## 8   RELATION TO EXISTING FRAMEWORKS

It is easy to see that our approach collapses to Prakken's strict argumentation framework (Prakken and Sartor, 1997) when only one knowledge base is present or the set of knowledge bases is non-disjunctive. Further, all arguments in rebuttal conflicts are *arguable* if priority is not available. In that situation, our framework reduces to partial semantics similar to Prakken's (Prakken and Sartor, 1997).

---

[2]See appendix for details

## 9   AN ILLUSTRATIVE EXAMPLE

A company would like to perform strategically planning and seek an answer to the following question: *Based on the current economic situations, is it profitable to start a new production line?*

To proceed, the company consulted two strategic planning experts, experts A and B. It was hoped that individual analysis from two independent sources could help review the intricacies of the scenario. Table 4 depicts knowledge set $KB_A$ and $KB_B$ extracted from expert A and B, respectively. The global order between experts is $A > B$.

Table 4: Knowledge Bases of Expert A and Expert B

| EXPERT | KNOWLEDGE |
|---|---|
| $KB_A$ | $A\_1 : \sim adversary\_financial\_factor \wedge economic\_grow \Rightarrow demand\_grow$ |
| | $A_2 : stable\_market \wedge demand\_grow \Rightarrow new\_production\_line \vee increase\_prod\_A$ |
| | $A_3 : \Rightarrow stable\_market$ |
| | $A_4 : \neg raw\_material\_A\_enough \Rightarrow \neg increase\_prod\_A$ |
| | $A_5 : \neg raw\_material\_A\_enough \Rightarrow \neg stable\_market$ |
| | $A_6 : \Rightarrow economic\_grow$ |
| | $A_7 : \Rightarrow \neg raw\_material\_A\_enough$ |
| | Preference Hierarchy $= \{A_3 > A_5\}$ |
| | |
| | $B_1 : \sim adversary\_financial\_factor \wedge economic\_grow \Rightarrow demand\_grow \vee competition\_grow$ |
| | $B_2 : competition\_grow \Rightarrow \neg stable\_market$ |
| | $B_3 : \sim \neg stable\_market \wedge \sim adversary\_financial\_factor \Rightarrow new\_production\_line$ |
| | $B_4 : \sim demand\_increase \Rightarrow \neg demand\_grow$ |
| | $B_5 : interest\_raise \Rightarrow adversary\_financial\_factor$ |
| | $B_6 : \Rightarrow interest\_raise$ |
| | $B_7 : \Rightarrow stock\_index\_raise$ |
| | $B_8 : stack\_index\_raise \Rightarrow \neg adversary\_financial\_factor$ |
| | $B_9 : \Rightarrow economic\_grow$ |
| | Preference Hierarchy $= \{B_8 > B_5\}$ |

In $KB_A$, $\{A_6, A_1, A_3, A_7, A_4, A_2\}$ form an argument $Arg_1$. The unique meaning of $Arg_1$ is *new_production_line* which suggests a new production line for the business. $\{A_7, A_5\}$ form another argument $Arg_2$. $Arg_2$ concludes that the market is stable which is contrary to the rule $A_3$ in $Arg_1$. By definition, $Arg_2$ rebut $Arg_1$. According to $KB_A$'s preference hierarchy, $Arg_1$ wins. Thus, $Arg_1$ is justified with respect to

$KB_A$.

In $KB_B$, $\{B_9, B_1, B_4, B_2\}$ form an argument $Arg_3$, $\{B_3\}$ form $Arg_4$, $\{B_6, B_5\}$ form $Arg_5$, $\{B_7, B_8\}$ form $Arg_6$. $Arg_3$ is undercut by $Arg_5$ which is then rebuted by $Arg_6$. $Arg_3$ undercuts $Arg_4$. According to the preference hierarchy, $Arg_6$ strictly defeats $Arg_5$ and $Arg_3$ is then justified. $Arg_4$ is strictly defeated by the justified argument $Arg_3$.

Consider the notion of skeptical view, we notice that $A_1$ of $KB_A$ and $B_1$ of $KB_B$ are similar knowledge about the relation among "adversary financial factor", "economic grow" and "demand grow". By definition, $B_1$ thins $arg_1$, the argument with $A_1$. The auxiliary rule is then,

$A_1^1 : \sim adversary\_financial\_factor \wedge economic\_grow$
$\Rightarrow competition\_grow$

To defeat $B_1$ and restore $Arg_1$'s justified status, we have to find arguments rebutting $A_1^1$. However, there is no such arguments. Thus, we have two different conclusions from $KB_A$ and $KB_B$. Credulous view suggests a new production line whereas skeptical view does not suggest any strategic moves. The result is close to our intuitive understanding of credulous reasoning and skeptical reasoning.

## 10    CONCLUSION

In this paper, we have discussed the inherent assumption in classical logic [3] and how it affects distributed reasoning under uncertainties. We have proposed an integrated framework for handling both intra-KBS-conflicts and inter-KBS-conflicts holistically. Furthermore, we have also discussed the general aspects of inter-KBS-conflicts for uncertainty reasoning.

As it is, our argumentation model is a little bit restrictive. It only works well in near horn (Reed et al., ) knowledge systems. Relaxing that restriction will lead us either to Stable semantics (Gelfond and Lifschitz, ) or Well-founded semantics (Ross, 1989). Such an extension not only bridges our framework to other frameworks in non-monotonic reasoning but also facilitates the modeling of more complex problems. In either way, it would be interesting to see how inter-KBS-conflicts are modeled and resolved. These lay down the core of our future work.

## Acknowledgements

This work is partially support by the direct grant from the Faculty of Engineering, CUHK, (grant number: 2050180). Thanks to invaluable comments from three anonymous referees.

---

[3]Certain information has higher preference than uncertain information.

# A APPENDIX

Before going into the soundness and completeness proof, we shall introduce the following notations to simplify the proof procedure. For an argument $Arg$,

- $tree(Arg)$ is the argument tree in which $Arg$ wins;
- $branch(T)$ is the set of branches of an argument tree $T$;
- $length(B)$ is the length of a branch;
- $move(B, n)$ is the $n$th move of a branch $B$;
- $player(M)$ is the player that responsible for the move $M$.

The proofs shown here are in *Lamport* style (Lamport, 1993).

**Monotone Proof**

**Lemma 1** Given a set of argument $S$ and conflict free subset $S_i$ and $S_{i+1} \subseteq S_i$. If an argument $Arg$ is in $\Pi_S(S_i)$ then $Arg$ is also in $\Pi_S(S_{i+1})$.

PROOF SKETCH: We prove the thesis by contradiction. We assume there is an argument $Arg$ in $\Pi(S_i)$ but not in $\Pi(S_{i+1})$. Using the definition of fix-point iteration, a contradiction is shown.
ASSUME: 1. ∃ argument $Arg$, s.t. $Arg \in \Pi(S_i)$ and $Arg \notin \Pi(S_{i+1})$
⟨1⟩1. $\exists DArg$ defeating $Arg$ which is not defeated by arguments in $S_{i+1}$.
⟨1⟩2. $S_i \subseteq S_{i+1}$ implies $DArg$ is also not defeated by $S_i$.
⟨1⟩3. $\exists DArg$ defeating $Arg$ which is not defeated by arguments in $S_i$.
⟨1⟩4. By definition, $Arg \in \Pi(S_i)$ implies $\not\exists DArg$ defeating $Arg$ which is not defeated by arguments in $S_i$.
□

**Theorem 1** The fix-point operator $\Pi$ is monotone.

PROOF: A direct rephrasing of lemma 1.

**Soundness Proof**

**Lemma 2** If arguments $Arg$ is provably justified, then there exists an argument tree such that every move of the proponent in every branch involves only justified arguments.

PROOF SKETCH: We prove the lemma by induction on the level of $tree(Arg)$. Let $h$ be the height of $tree(Arg)$. We argue that all proponent moves at level $h$ are justified. By backward induction, we argue that proponent moves at odd level $i$ must also be justified based on justified proponent moves at level $i + 2$.
ASSUME: 1. ∃ $tree(Arg)$ s.t. $Arg$ wins all branches.
    2. $n = length(tree(Arg))$
    3. $level(Arg, i) = \{m | m \in move(\ branch(\ tree(Arg)\ ), i)\ \}$
⟨1⟩1. **Basis** : at level $h = n$, $\forall m \in level(Arg, h)$, $m$ is justified
    ⟨2⟩1. $player(m) = proponent$
    ⟨2⟩2. $m$ is the last move implies $\not\exists \bar{m}$ defeating $m$
    ⟨2⟩3. $m$ is justified by definition
    ⟨2⟩4. Q.E.D.
⟨1⟩2. **Induction Step** : If $i$ is odd and $level(Arg, i+2)$ is justified, then $level(Arg, i)$ is also justified
    ASSUME: 1. $level(Arg, i+2)$ is justified.
        2. $leaf = \{m | m \in level(Arg, i)$ and $m$ is not the last move of the branch $\}$
        3. $non\text{-}leaf = level(Arg, i) - leaf$
    ⟨2⟩1. $leaf$ is justified, by the same reason in ⟨1⟩1.
    ⟨2⟩2. $\forall m \in non\text{-}leaf, \exists \bar{m} \in level(Arg, i+1)$ strictly defeat it.
    ⟨2⟩3. since $m \in non\text{-}leaf$ and end move of every branch is by proponent, there exists a move $m^*$ by proponent after $\bar{m}$.
    ⟨2⟩4. $m^*$ is justified as $m^*$ in $level(Arg, i+2)$.
    ⟨2⟩5. $\forall \bar{m}$ of $m$, $\bar{m}$ is strictly defeated.
    ⟨2⟩6. $m$ is justified by fix-point definition.
    ⟨2⟩7. Q.E.D.

⟨1⟩3. **Completion** : For all odd integer $i < n$, $level(Arg, i)$ is justified. Moves of proponent only occur at odd level of an argument tree.
□

**Theorem 2** All provably justified arguments are justified.

PROOF SKETCH: Every provably justified argument, $Arg$, has an argument tree in which all proponent moves are justified. $Arg$ is one of those proponent moves and the result follows.
⟨1⟩1. By lemma 2, $\exists t \in tree(Arg)$ s.t. all proponent moves of $t$ are justified.
⟨1⟩2. Root of $t$ is justified.
⟨1⟩3. $Arg$ is root of $t$.
⟨1⟩4. $Arg$ is justified.
□

**Completeness Proof**

**Lemma 3** If $Arg$ is a justified argument, then there exists an argument tree such that every moves of the proponent involves only justified arguments.

PROOF SKETCH: We start to construct a tree inductively with a justified argument $Arg$. At every odd level $i$, we show that there exists justified arguments $DArg$ strictly defeating the defeaters of level $i$. $DArg$ can then form level $i + 2$'s move. Then, we can inductively construct an argument tree by non repetitive moves of the opponent.
ASSUME: 1. $Arg$ is justified.
    2. $T$ is an argument tree.
    3. $Arg$ is at level 1 of $T$.
⟨1⟩1. **Basis** : Level 3 is justified.
    ASSUME: 1. $D$ is the set of defeaters defeating $level(Arg, 1)$.
    ⟨2⟩1. $\forall d \in D$, $\exists \hat{d} \in$ fix-point strictly defeating $d$.
    ⟨2⟩2. $DArg = \{\hat{d} | d \in D\}$ is justified by definition.
    ⟨2⟩3. let $level(Arg, 3) = DArg$
    ⟨2⟩4. Q.E.D.
⟨1⟩2. **Induction Step** : If level $i$ is justified, there exists justified $level(Arg, i+2)$ defeating $level(Arg, i+1)$
    ASSUME: 1. $level(Arg, i)$ is justified.
        2. $level(Arg, i+1)$ is non-empty.
        3. $arguable = \{m | m \in level(Arg, i) \cap m$ is defeated by $level(Arg, i+1)\ \}$
    ⟨2⟩1. $level(Arg, i+1)$ defeats $level(Arg, i)$
    ⟨2⟩2. $\forall m \in arguable$, there exists $\bar{m}$ in fix-point defeating $level(Arg, i+1)$

    ⟨2⟩3. $level(Arg, i+2) = \{\bar{m} | m \in arguable\ \}$
    ⟨2⟩4. $level(Arg, i+2)$ is justified and defeats $level(Arg, i+1)$
    ⟨2⟩5. Q.E.D.
⟨1⟩3. **Completion** : Thus, there exists a proof tree in which every proponent level involves only justified arguments.
□

**Theorem 3** All justified arguments are provable.

PROOF SKETCH: We prove the thesis by contradiction. Start with there exists a justified argument $Arg$ that is not provable. Using lemma 3 we find an argument tree to support $Arg$ and lead to a contradiction.
ASSUME: 1. $Arg$ is justified.
    2. By Lemma 3 and the hypothesis, $\exists T$, an argument tree such that
        a. every move of a proponent is justified
        b. a proponent cannot move in a branch $B$
⟨1⟩1. Let
    1. $m = move(B, length(B))$
    2. $\bar{m} = move(B, length(B)\text{-}1)$
⟨1⟩2. $\bar{m}$ is justified.
⟨1⟩3. $\exists m'$ , s.t. $m'$ strictly defeats $m$.
⟨1⟩4. $m'$ is a viable move for the proponent.
□